\documentclass{llncs}

\usepackage{url}
\usepackage{graphicx}
\usepackage{booktabs}
\usepackage{amsmath}
\usepackage{amssymb}
\usepackage{xcolor}
\usepackage{siunitx}
\usepackage{microtype}
\usepackage{float}
\usepackage{placeins}
\usepackage{needspace}

\AtBeginDocument{\renewcommand{\doi}[1]{\url{https://doi.org/#1}}}

\DeclareSIUnit\mebibyte{MiB}
\newcommand{\FF}{\texttt{FF}}
\newcommand{\MF}{\texttt{MF}}
\newcommand{\BP}{\texttt{BP}}
\newcommand{\BPDS}{\texttt{BP-DS}}
\newcommand{\MFJ}{\texttt{MF-Joint}}
\newcommand{\CaFo}{\texttt{CaFo}}

\title{Energy-Efficient Deep Learning\\ Without Backpropagation: A~Rigorous\\
Hardware-Validated Benchmarking Study\\ of Forward-Only Algorithms}

\author{Przemys{\l}aw Spyra\orcidID{0009-0009-3001-2604} \and
Witold~Dzwinel\orcidID{0000-0001-8321-5928}}
\titlerunning{Energy-Efficient Deep Learning Without Backpropagation}
\authorrunning{P. Spyra and W. Dzwinel}

\institute{AGH University of Krakow, Faculty of Computer Science,
Krak\'ow, Poland\\
\email{pspyra@agh.edu.pl, dzwinel@agh.edu.pl}}

\makeatletter
\renewcommand{\@cite}[2]{\mbox{[{#1\if@tempswa , #2\fi}]}}
\makeatother

\begin{document}
\maketitle

\begin{abstract}
Backpropagation (\BP{}) powers modern deep learning, yet it is biologically
implausible and computationally awkward: every activation must stay in memory
until a global backward sweep consumes it, and that sweep locks each layer
until the one above has finished. Forward-only algorithms, from Hinton's
Forward-Forward (\FF{}) through Cascaded Forward (\CaFo{}) to Mono-Forward
(\MF{}), replace the global gradient with local learning signals and are promoted as
more efficient on the basis of FLOP counts rather than measurements. We test that claim on hardware. Across almost \num{1000} runs on four
vision benchmarks (MNIST, Fashion-MNIST, CIFAR-10 and CIFAR-100), we read energy, power and memory from an NVIDIA
A100 GPU and compared each algorithm against a tuned \BP{} baseline. Removing
the backward pass alone saves nothing: \FF{} costs \SIrange{4.1}{6.3}{\times}
\BP{}'s energy and \CaFo{} \SIrange{3.0}{7.1}{\times}. Mono-Forward is different. Given the same number of data passes
as \BP{}, it uses \SIrange{12.5}{19.9}{\percent} less energy and up to
\SI{27.5}{\percent} less memory on the \num{3}$\times$\num{2000} multilayer perceptrons, the
largest configurations tested, at lower power in comparable time; on
CIFAR-100 its test accuracy is statistically equivalent to \BP{}'s. A
pre-registered ablation identifies gradient detachment, not \MF{}'s auxiliary
losses, as the dominant factor tested. Trained to convergence, \MF{} exceeds
\BP{}'s test accuracy on CIFAR-10 and CIFAR-100 by \num{3.90} and
\num{5.04} percentage points, with \SIrange{3.5}{4.5}{\times} as many passes. A
forward-only algorithm can thus match backpropagation's accuracy at lower
energy and, given more passes, surpass it; faster convergence is the remaining target
for an end-to-end gain.

\keywords{\def\and{\unskip\nobreak\ \textperiodcentered\nobreak\ \ignorespaces}%
\hyphenpenalty=10000 \exhyphenpenalty=10000
Backpropagation-free learning \and Energy efficiency \and
Hardware benchmarking \and Forward-only algorithms \and Deep neural networks}
\end{abstract}

\section{Introduction}
\label{sec:intro}

The energy cost of training deep neural network (DNN) models is no longer negligible.
Training footprints have grown faster than the efficiency of the
underlying hardware~\cite{strubell2019energy,patterson2021carbon}, and
the gap between what accelerators deliver and what workloads actually draw
is now a first-order engineering concern~\cite{jouppi2017datacenter,wef2025paradox}.
What the training algorithm itself costs is therefore worth measuring directly.
Backpropagation (\BP{})~\cite{rumelhart1986learning} sits at the centre of that
cost. It requires every intermediate activation of the forward pass to remain
live until the backward pass consumes it, which fixes a memory floor
proportional to depth and width, and it serialises the backward sweep, which
fixes a dependency chain that no amount of parallel hardware
removes~\cite{jaderberg2017decoupled,lillicrap2020backpropagation}.

Forward-only algorithms remove the backward sweep. Forward-Forward
(\FF{})~\cite{hinton2022forward}, Cascaded Forward
(\CaFo{})~\cite{zhao2023cafo} and Mono-Forward
(\MF{})~\cite{gong2025mono} each replace the global gradient with signals
computed locally, and each is proposed, at least in part, on efficiency
grounds. The evidence offered for that efficiency is usually analytical: a
count of floating-point operations (FLOPs), a parameter count, an argument about what need not be stored. Very
little of it is measured on hardware, and an analytical proxy cannot see the
two things that dominate a real training run, which are how much power the
device draws while it works and how many passes over the data the algorithm
needs before it stops.

This paper measures both. We instrument every run through the NVIDIA
Management Library (NVML), the driver's own telemetry interface, and read
power, utilisation, clock and memory from the device at \SI{5}{\hertz} while
the workload executes. Across about \num{1000} runs on four vision benchmarks
(MNIST, Fashion-MNIST, CIFAR-10 and CIFAR-100), and given the same number of
data passes as \BP{}, Mono-Forward uses \SIrange{12.5}{19.9}{\percent} less energy and
\SIrange{24.6}{27.5}{\percent} less allocator memory than backpropagation on
the multilayer perceptrons (MLPs) with three hidden layers of \num{2000} units
(\num{3}$\times$\num{2000}) on CIFAR-10 and CIFAR-100, the largest
configurations tested; on the \num{2}$\times$\num{1000} MLPs trained on MNIST
and Fashion-MNIST the energy saving is \SIrange{2.2}{4.3}{\percent}. \MF{} is
cheaper because it draws less power, not because it runs faster, so the
comparison does not depend on when a run was allowed to stop.

Under a fixed budget the cheaper pass appears as lower cost. Trained to its
own convergence, \MF{} reaches test accuracy above \BP{}'s by
\num{3.90} and \num{5.04} percentage points (pp) on CIFAR-10 and CIFAR-100,
at the price of more passes.

This paper makes four contributions. The first is the matched data-pass
protocol, a comparison in which \MF{} receives at least as many data passes as \BP{}
actually used, so the stopping rule cannot confound the result; under it \MF{}
uses \SIrange{12.5}{19.9}{\percent} less energy and
\SIrange{24.6}{27.5}{\percent} less allocator memory on the
\num{3}$\times$\num{2000} MLPs (Section~\ref{sec:matched},
Table~\ref{tab:matched}). The second is that the advantage is
a \emph{power} advantage at near-equal wall-clock time and that it is larger in
the larger configurations tested, from no memory advantage at \num{1.8}~M
weights to \SI{27.5}{\percent} at \num{14.2}~M (Figure~\ref{fig:width}). The
third is a pre-registered ablation ladder that identifies gradient detachment,
rather than the localised objectives, as the dominant factor among those
tested: a deep supervision control is \emph{worse} than plain \BP{} on both
CIFAR tasks, while \MF{} exceeds it by \num{5.56} and \num{6.10}~pp in test
accuracy (Section~\ref{sec:ladder}). The fourth observation is
that the convergence rate is the only remaining barrier to \MF{} achieving an
end-to-end energy advantage. This rate, quantified as a
\SI{3.5}{\times}--\SI{4.5}{\times} pass multiplier, defines the target that
schedule design should aim to meet (Section~\ref{sec:limits}).

\section{Background and Related Work}
\label{sec:background}

\paragraph{The cost of the backward pass.}
\BP{} computes exact gradients by storing the forward activations and
traversing the graph in reverse. The consequences are structural: activation
memory scales with depth, width and batch size, and the backward traversal
cannot begin until the forward pass ends (forward locking) or be split across
layers without approximation (backward locking)~\cite{jaderberg2017decoupled}.
Alternatives include predictive coding~\cite{pinchetti2022predictive,salvatori2022brain},
Hebbian and associative schemes~\cite{journe2023hebbian,salvatori2022associative,salvatori2024stable},
and zeroth-order methods for large-model
fine-tuning~\cite{malladi2023fine}; Lillicrap et al.~\cite{lillicrap2020backpropagation}
survey the space that the biological case against \BP{} opens: the symmetric feedback weights and
separate backward phase that \BP{} requires are difficult to reconcile with
what is known about cortical circuits.

\paragraph{Forward-only and local-loss methods.}
We write vectors in bold lower case and matrices in bold upper case. A network
with hidden layers $i = 1, \ldots, L$ has a weight matrix $\mathbf{W}_i$ per
layer and activation vectors $\mathbf{a}_i = f(\mathbf{W}_i \mathbf{a}_{i-1})$,
where $\mathbf{a}_0 = \mathbf{x}$ is the input, $f$ is the activation function
and biases are omitted. Forward-Forward (\FF{})~\cite{hinton2022forward}
trains each layer to separate positive (real) from negative (corrupted) data
by a goodness score, the sum of its squared activations, so no gradient
crosses a layer boundary. Cascaded Forward (\CaFo{})~\cite{zhao2023cafo}
attaches a predictor to every block of a convolutional cascade and trains each
predictor independently of the others; a variant first pre-trains the blocks
with direct feedback alignment (DFA). Mono-Forward (\MF{})~\cite{gong2025mono}
attaches to each activation $\mathbf{a}_i$ a learned projection matrix $\mathbf{M}_i$ that maps it to
class logits $\mathbf{M}_i \mathbf{a}_i$, trains each layer on a local
cross-entropy loss over those logits, and trains the layers one after another.
Related local-objective work has scaled layer-wise training to larger
settings~\cite{ren2023scaling} and analysed it in the infinite-width
limit~\cite{ishikawa2025local}, where greedy layer-wise optimisation reduces
the effective complexity of the learned representation.

\paragraph{Deep supervision.}
Auxiliary classifiers attached to intermediate layers predate the forward-only
literature~\cite{lee2015deeply} and are the obvious control for it: they supply
the same local supervisory signal while leaving the global gradient intact,
and layer-wise objectives have been studied as a training device in their own
right~\cite{lorberbom2024layer}. Section~\ref{sec:ladder} applies that control.

\paragraph{Hardware and energy.}
Analytical proxies systematically misestimate real
consumption~\cite{desislavov2024energy,abbas2024impact}, because they cannot
see clock behaviour, utilisation or the fixed costs of a kernel launch.
Direct device telemetry is the alternative, and recent forward-only work has
begun to adopt it~\cite{papachristodoulou2024convolutional}. We use the NVIDIA
Management Library (NVML) throughout and name the instrument behind every
quantity.

\needspace{8\baselineskip}
\section{Mono-Forward and the Cost Hypothesis}
\label{sec:mf}

\MF{} trains stage $i$ of an MLP by minimising the cross-entropy of $\mathbf{M}_i \mathbf{a}_i$ against the
true class label, so that each layer learns to classify the input on its own. Gradients of this loss
are computed with respect to $\mathbf{W}_i$ and $\mathbf{M}_i$ only: the input
$\mathbf{a}_{i-1}$ is treated as a constant, detached from the computational
graph, so no gradient propagates to earlier layers. Training proceeds one
stage at a time, so a network with $L$ hidden layers has $L+1$ stages, one per
projection including $\mathbf{M}_0$ on the raw input. At inference the network
is a standard feedforward stack with $\mathbf{M}_L$ as the readout.

Three consequences follow, and they are the reason to expect a cost advantage.
No activation needs to outlive its stage, so the allocator high-water mark is
set by the widest single layer rather than by the whole forward pass. Each
optimiser step updates one weight matrix and one projection, so the arithmetic
per step is smaller and the kernels narrower. Finally, every layer below the
current stage is frozen throughout that stage, so its activations depend only
on the sample and can be cached, which no jointly trained network permits.
The countervailing effect is equally structural: stage $i$ must reconstruct
$\mathbf{a}_{i-1}$ before it can train, which yields an $O(L^2)$ total when
recomputation is used. Our experiments recompute, and
Section~\ref{sec:objections} evaluates whether caching is beneficial. \MF{}
spends less work per step and more per epoch, so the sign of the net
effect is an empirical question about a device, and the rest of this paper answers it.

\section{Measuring Cost Fairly: Two Protocols}
\label{sec:protocols}

Energy and time are integrals over the duration of training, so whatever
decides when training stops also shapes the efficiency comparison. Every result below uses one
stopping rule, applied identically to every algorithm, and we report two
protocols rather than one, because no single protocol answers both of the
questions a practitioner asks: what an algorithm costs at its best, and what it costs for the same number of data passes.

\paragraph{Protocol A: own convergence.}
Every algorithm stops on validation loss with the same minimum improvement
(\num{0}), the same patience (\num{20}) and the same cap, an upper bound of
\num{100} epochs on each training unit (the whole network for \BP{}, each stage
for \MF{}). Each method therefore trains until it stops improving. This answers ``what does it
cost to get the best this method can do'', and it is the protocol under which
\MF{}'s accuracy advantage appears.

\paragraph{Protocol B: matched data-pass protocol.}
\MF{} is given the number of data passes \BP{} actually spent, and nothing
else changes. The budget is
\begin{equation}
  E_{\MF{}}^{\text{per stage}}
  \;=\; \left\lceil \frac{\bar{E}_{\BP{}}}{S} \right\rceil ,
  \qquad
  E_{\MF{}}^{\text{total}} \;=\; S \cdot E_{\MF{}}^{\text{per stage}},
  \label{eq:budget}
\end{equation}
where $\bar{E}_{\BP{}}$ is \BP{}'s mean epoch count under Protocol A
and $S$ is \MF{}'s stage count.

Rounding up with the ceiling function
$\lceil\cdot\rceil$, which is unrelated to the epoch cap of Protocol A, is
deliberate: it can only give \MF{} \emph{more} passes than \BP{}
had, so a saving measured under this protocol cannot be an
artefact of a smaller budget. Table~\ref{tab:budget} gives the resulting epoch budgets.
\BP{} itself is not rerun under Protocol B: it keeps its own early stopping and is
never pushed past the point it chose to stop, which is why the \BP{} column is
the same in both protocols. The budget equalises data passes, not optimiser
work per layer: because \MF{} trains one stage at a time, each \MF{} layer
receives $\lceil \bar{E}_{\BP{}}/S \rceil$ updates where each \BP{} layer
receives $\bar{E}_{\BP{}}$, which is part of why the budget costs \MF{}
accuracy on CIFAR-10.

Early stopping stays
\emph{enabled} with patience equal to the cap, so it cannot fire inside the
budget, and \MF{} still runs the per-epoch validation pass that \BP{} runs.

\begin{table}[tb]
  \centering
  \caption{The matched data-pass budget of Equation~\ref{eq:budget}, with
    \BP{}'s realised mean epoch count measured under Protocol A.}
  \label{tab:budget}
  \setlength{\tabcolsep}{8pt}
  \footnotesize
  \begin{tabular}{lrrrr}
    \toprule
    Configuration & \BP{} mean & \MF{} & Per & \MF{} \\
                  & epochs     & stages & stage & total \\
    \midrule
    MNIST 2$\times$1000     & 25.4 & 3 &  9 & 27 \\
    Fashion 2$\times$1000   & 25.6 & 3 &  9 & 27 \\
    CIFAR-10 3$\times$2000  & 52.0 & 4 & 13 & 52 \\
    CIFAR-100 3$\times$2000 & 42.9 & 4 & 11 & 44 \\
    \bottomrule
  \end{tabular}
\end{table}

\paragraph{What else is held fixed.}
Each algorithm is run on the architecture specified in its source publication.
For the MLP experiments, CIFAR inputs are flattened following the \MF{}
protocol~\cite{gong2025mono}. Each \BP{} baseline uses the same module list,
with only the algorithm-specific components removed.
Optuna~\cite{akiba2019optuna} tuned every algorithm and baseline under an
identical trial budget and epoch cap, unpruned, with the sampler seed derived
from the study name; \BP{} and \MF{} search two hyperparameters (learning rate and
weight decay), the ladder controls \BPDS{} and \MFJ{} of Section~\ref{sec:ladder} three (adding the auxiliary-loss weight), all
at \num{50} trials. All MLPs use ReLU and a batch size of \num{128}, with Kaiming
uniform initialisation~\cite{he2015delving} for every weight and projection;
\BP{} and \BPDS{} use AdamW~\cite{loshchilov2019decoupled}, \MFJ{} and \MF{}
stages use Adam~\cite{kingma2015adam} on a local cross-entropy. All algorithms share one PyTorch codebase and the
same \texttt{nn.Linear} primitives, differing only in the loss and in where
\texttt{detach()} is called, so no method gains a fused kernel the others lack.

\paragraph{Instruments.}
A monitor thread running alongside each job queries the
NVIDIA Management Library (NVML) at \SI{5}{\hertz} for power
(\texttt{nvmlDeviceGetPowerUsage}), utilisation
(\texttt{nvmlDeviceGetUtilizationRates}), streaming multiprocessor clock
(\texttt{nvmlDeviceGetClockInfo}) and device memory
(\texttt{nvmlDeviceGetMemoryInfo}); energy in \si{\watt\hour} is the
trapezoidal integral of power over the run. Wall-clock time comes from the
monotonic clock of the host.
The NVML memory reading covers the whole device and includes a fixed
${\approx}\SI{870}{\mebibyte}$ CUDA context that an algorithm can
neither reduce nor control. Every memory number in this paper is instead the
per-process allocator high-water mark,
\texttt{torch.cuda.max\_memory\_allocated}.

\paragraph{Setup and statistics.}
Runs used NVIDIA A100-SXM4-40GB graphics processing units (GPUs) on AMD EPYC
7742 nodes of the Athena cluster at ACK Cyfronet AGH, driver 595.71.05, Python~3.10.4, PyTorch~2.4.0+cu121. CIFAR runs
used random horizontal flips and \num{4}-pixel-padded random crops, applied
identically to every algorithm and its baseline. MNIST and Fashion-MNIST were
not augmented. Accuracy is measured on the held-out test split, and results are
mean~$\pm$~standard deviation (SD) over the seeds stated per table.
\emph{The ablation ladder of Section~\ref{sec:ladder} is pre-registered}: the
analysis script fixing the paired-by-seed primary test, the Holm-Bonferroni
correction within each metric family, the \num{10000}-resample bootstrap
intervals and the $\pm\num{0.25}$~pp equivalence margin was written and
committed before any ladder result was inspected. Parity claims require the
two one-sided tests (TOST) procedure~\cite{schuirmann1987comparison,lakens2017equivalence},
which declares two arms equivalent only when the \num{90}\% confidence
interval (CI) of their difference lies inside the margin. A contrast is EQUIVALENT
when TOST succeeds at $\alpha = 0.05$, otherwise DIFFERENT when the difference
test rejects, and otherwise INCONCLUSIVE.
The matched data-pass protocol of Section~\ref{sec:matched} was
\emph{not} pre-registered; it is a diagnostic, analysed with the same
machinery and margins, not a confirmatory test.

\section{Results}
\label{sec:results}

\subsection{The Matched Data-Pass Protocol: MF Is the Cheaper Algorithm}
\label{sec:matched}

Table~\ref{tab:matched} summarises the main experimental results. Given the budget of
Equation~\ref{eq:budget}, \MF{} draws \SI{19.9}{\percent} less GPU energy than
\BP{} on the CIFAR-10 \num{3}$\times$\num{2000} MLP and \SI{12.5}{\percent}
less on CIFAR-100, holds \SI{27.5}{\percent} and \SI{24.6}{\percent} less
per-process allocator memory, and runs at \SI{63.3}{\watt} and \SI{62.8}{\watt}
of mean board power against \BP{}'s \SI{76.8}{\watt} and \SI{72.0}{\watt}.
All of these contrasts are DIFFERENT at $n=20$ paired seeds, and
on CIFAR-100 accuracy is statistically equivalent to \BP{}'s. Three properties of these measurements make the result stronger than the
percentages suggest.

\emph{The memory figures are exact.} Peak allocator occupancy has zero
seed-to-seed standard deviation in every arm of every configuration, because
the allocation schedule of a fixed architecture at a fixed batch size is
deterministic, so the \SI{82.2}{\mebibyte} and \SI{74.4}{\mebibyte} that \MF{}
saves need no confidence interval.

\emph{The energy saving is a power saving.} Wall-clock time under the matched
data-pass protocol is not reduced: in the column order of Table~\ref{tab:matched} the
four contrasts give $+3.8$, $-0.8$, $-2.8$ and
$+0.4\,\%$, of which CIFAR-100 is EQUIVALENT within a
$\pm\SI{5}{\percent}$ margin, MNIST and CIFAR-10 are INCONCLUSIVE, and
Fashion-MNIST is DIFFERENT with \MF{} the faster.
Power is a rate, so no stopping rule can influence it,
though at this utilisation it is specific to this device
(Section~\ref{sec:discussion}).

\emph{On the most demanding task the saving costs no accuracy.} On CIFAR-100,
with \num{100} classes, \MF{}
reaches accuracy statistically equivalent to \BP{}'s under the budget
($+\num{0.02}$~pp, 90\% CI $[\num{-0.13}, \num{+0.16}]$, TOST EQUIVALENT) while
spending \SI{12.5}{\percent} less energy and \SI{24.6}{\percent} less memory.
Elsewhere the budget costs \MF{}
some accuracy, \num{1.11}~pp on CIFAR-10 $[\num{-1.32}, \num{-0.91}]$ and
\num{0.22} and \num{0.49}~pp on the two \num{2}$\times$\num{1000} tasks, as
expected of a method stopped short of convergence
(Section~\ref{sec:convergence}); the energy saving points the same way on all
four and is resolved on three.

\begin{table}[tb]
  \centering
  \caption{Matched data-pass protocol (Protocol B). \MF{} receives the budget of
    Table~\ref{tab:budget}; \BP{} is its own Protocol A run. Energy is the
    trapezoidal integral of the \SI{5}{\hertz} NVML power trace; memory is
    \texttt{torch.cuda.max\_memory\_allocated}, per process; power is the trace
    mean. $\Delta$ rows are \MF{} relative to \BP{}, so negative is cheaper.
    \MF{} arms have $n=20$; \BP{} arms have
    $n=\num{60}, \num{122}, \num{20}, \num{20}$ by column. Memory has zero
    seed-to-seed SD.}
  \label{tab:matched}
  \setlength{\tabcolsep}{5pt}
  \footnotesize
  \begin{tabular}{lrrrr}
    \toprule
    & MNIST & Fashion & CIFAR-10 & CIFAR-100 \\
    & 2$\times$1000 & 2$\times$1000 & 3$\times$2000 & 3$\times$2000 \\
    \midrule
    \MF{} epoch budget & $27$ & $27$ & $52$ & $44$ \\
    \addlinespace[2pt]
    \multicolumn{5}{l}{\emph{Test accuracy} (\%)} \\
    \quad \BP{} & $98.20$ & $89.41$ & $62.47$ & $32.90$ \\
    \quad \MF{} & $97.98$ & $88.92$ & $61.36$ & $32.92$ \\
    \quad $\Delta$ (pp) & $-0.22$ & $-0.49$ & $-1.11$ & $+0.02$ \\
    \addlinespace[2pt]
    \multicolumn{5}{l}{\emph{GPU energy} (\si{\watt\hour})} \\
    \quad \BP{} & $1.079$ & $1.100$ & $5.393$ & $4.141$ \\
    \quad \MF{} & $1.055$ & $1.053$ & $4.320$ & $3.624$ \\
    \quad $\Delta$ (\%) & $-2.2$ & $-4.3$ & $-19.9$ & $-12.5$ \\
    \addlinespace[2pt]
    \multicolumn{5}{l}{\emph{Peak allocated memory} (\si{\mebibyte}, per process)} \\
    \quad \BP{} & $50.9$ & $50.9$ & $299.0$ & $302.5$ \\
    \quad \MF{} & $51.0$ & $51.0$ & $216.8$ & $228.1$ \\
    \quad $\Delta$ (\%) & $+0.3$ & $+0.3$ & $-27.5$ & $-24.6$ \\
    \addlinespace[2pt]
    \multicolumn{5}{l}{\emph{Mean board power} (\si{\watt})} \\
    \quad \BP{} & $62.9$ & $61.5$ & $76.8$ & $72.0$ \\
    \quad \MF{} & $59.3$ & $59.3$ & $63.3$ & $62.8$ \\
    \quad $\Delta$ (\%) & $-5.8$ & $-3.6$ & $-17.6$ & $-12.8$ \\
    \addlinespace[2pt]
    \emph{Wall-clock time} $\Delta$ (\%) & $+3.8$ & $-0.8$ & $-2.8$ & $+0.4$ \\
    \bottomrule
  \end{tabular}
\end{table}

\subsection{The Advantage Is Larger in the Larger Configurations}
\label{sec:width}

\begin{figure}[tb]
  \centering
  \includegraphics{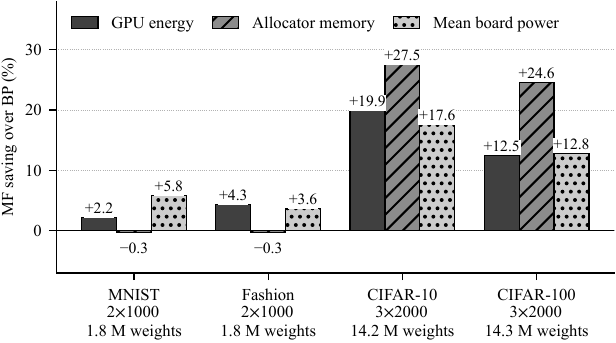}
  \vspace{-3pt}
  \caption{Cost saving of \MF{} over \BP{} under the matched data-pass protocol,
    ordered by trainable weights. Positive is cheaper. Every
    axis is larger at the \num{3}$\times$\num{2000} size, though not
    monotone within a size; the memory advantage only appears at the larger
    size, where the \num{0.14}~\si{\mebibyte} of extra projection parameters
    stops dominating. Bars are seed means, $n=20$ for \MF{} and
    \numrange{20}{122} for \BP{}. Memory is deterministic within an arm.}
  \label{fig:width}
\end{figure}

The four configurations span a nearly eightfold range of trainable weights, from
\num{1.79}~M on the \num{2}$\times$\num{1000} MLPs to \num{14.16} and
\num{14.34}~M on the \num{3}$\times$\num{2000} MLPs, and every cost axis moves
the same way across that range (Figure~\ref{fig:width}). Energy saving rises from
\SI{2.2}{\percent} and \SI{4.3}{\percent} on the smaller configurations to
\SI{12.5}{\percent} and \SI{19.9}{\percent} on the larger; power saving from
\SI{3.6}{\percent} and \SI{5.8}{\percent} to \SI{12.8}{\percent} and
\SI{17.6}{\percent}; and memory from a \SI{0.3}{\percent} \emph{deficit}, which
is the \SI{0.14}{\mebibyte} of extra projection parameters, to a saving of
\SI{24.6}{\percent} and \SI{27.5}{\percent}.

The mechanism is visible in the utilisation trace. Over these four
configurations \BP{} holds the device at \SI{11.5}{\percent} mean utilisation
and never exceeds \SI{17.3}{\percent}, while \MF{} averages \SI{6.0}{\percent}
and never exceeds \SI{8.0}{\percent}. Both are far from saturation: the device
idles between kernels for most of each run, and \MF{} keeps it busy about half
as long. We read the widening gap as an effect of the narrower per-stage
kernels, which the trace is consistent with but does not prove.

The four configurations give only two width levels, and width moves together
with depth and task across them, so Figure~\ref{fig:width} cannot separate
width from scale. It does show that the advantage is not a small-model
artefact: \MF{} does better on the larger networks.

\subsection{Where the Accuracy Comes From}
\label{sec:ladder}

\begin{figure}[tb]
  \centering
  \includegraphics{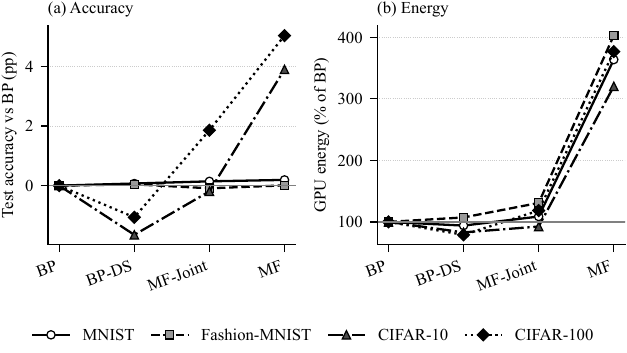}
  \vspace{-3pt}
  \caption{The ablation ladder, one structural change per step. (a) Test accuracy against \BP{}. Deep
    supervision (\BPDS{}) falls below \BP{} on both CIFAR tasks; the readout
    step recovers most or all of that, and the detach is the largest single
    gain on both. (b) GPU energy as a percentage of \BP{}.
    The same step carries essentially the whole cost increase. Markers are seed
    means over \numrange{20}{122} seeds per rung.}
  \label{fig:ladder}
\end{figure}

\MF{} differs from \BP{} in three ways at once: it adds auxiliary losses, it
adds the projections $\mathbf{M}_i$, and it detaches the gradient. We therefore
interpolate between \BP{} and \MF{} in single structural steps and measure each (Table~\ref{tab:ladder}, Figure~\ref{fig:ladder}):

\begin{center}\footnotesize
\setlength{\tabcolsep}{6pt}
\begin{tabular}{llcll}
  \toprule
  Rung    & Readout       & Auxiliary losses & Gradients & Optimiser \\
  \midrule
  \BP{}   & output layer  & no  & global & AdamW \\
  \BPDS{} & output layer  & yes & global & AdamW \\
  \MFJ{}  & $\mathbf{M}_L$ & yes & global & Adam \\
  \MF{}   & $\mathbf{M}_L$ & yes & \emph{local} & Adam \\
  \bottomrule
\end{tabular}
\end{center}

Each rung is tuned independently, so every step also moves the learning rate
and weight decay, and \BPDS{}$\rightarrow$\MFJ{} crosses AdamW to Adam: that
step is not attributable to the readout alone. The step the mechanism rests on,
\MFJ{}$\rightarrow$\MF{}, holds the optimiser family fixed.
\BPDS{} is the standard deep supervision control: \MF{}'s
auxiliary classifiers attached to \BP{} with nothing detached.

On the two CIFAR \num{3}$\times$\num{2000} MLPs, deep supervision on its own is
\emph{worse} than plain \BP{}, by \num{1.65}~pp
$[\num{-1.85}, \num{-1.45}]$ on CIFAR-10 and \num{1.06}~pp
$[\num{-1.26}, \num{-0.87}]$ on CIFAR-100. \MF{} beats that control (Table~\ref{tab:ladder}) by
\num{5.56}~pp $[\num{+5.37}, \num{+5.75}]$ and \num{6.10}~pp
$[\num{+5.88}, \num{+6.32}]$, these two contrasts at Holm-corrected $p < 10^{-20}$ and
the two above at $p < 10^{-8}$; all are \num{95}\% bootstrap
intervals over \num{20} paired seeds. Auxiliary supervision therefore does
not explain the accuracy; added to \BP{}, it lowers \BP{}'s accuracy. Among the
factors tested, detaching the gradient is the largest single contributor: the
\MFJ{}$\rightarrow$\MF{} step, which detaches it with the optimiser family
held fixed, moves accuracy by
$+\num{4.10}$ and $+\num{3.18}$~pp on CIFAR-10 and CIFAR-100.

On the two \num{2}$\times$\num{1000} MLPs the same comparison is a tie: \MF{}
leads \BPDS{} by \num{0.11}~pp $[\num{-0.01}, \num{+0.23}]$ on MNIST and trails
by \num{0.06}~pp $[\num{-0.17}, \num{+0.06}]$ on Fashion-MNIST, both EQUIVALENT
within the pre-registered $\pm\num{0.25}$~pp equivalence margin. We report both halves of the answer: on the small tasks
the experiment does not distinguish an \MF{}-specific benefit from auxiliary
supervision, and on the harder ones auxiliary supervision lowers accuracy and
the detach is the largest single step.

In energy, the detach step is also the expensive one: it adds
\numrange{228}{272}~pp of \BP{}'s energy across the four configurations, while
auxiliary supervision and the readout together move energy by at most
\num{31}~pp, and on CIFAR-10 move it in the opposite direction
($-\num{7.5}$~pp). The detach step is thus the largest contributor to both
the accuracy gain and the cost, which is why the two protocols disagree and
why neither alone is a complete answer.

\begin{table}[tb]
  \centering
  \caption{The \BP{}$\rightarrow$\MF{} ablation ladder under Protocol A, one
    structural change per step; rungs are tuned independently, so
    \BPDS{}$\rightarrow$\MFJ{} also crosses AdamW to Adam. $\Delta$ is accuracy
    against \BP{} in pp, as a difference of arm means; where arms have unequal
    $n$ the paired contrast differs (MNIST, $+\num{0.25}$~pp). $E$ is GPU
    energy as a percentage of \BP{}'s. Seeds per rung: \numrange{20}{122}. This analysis is pre-registered; intervals and
    Holm-corrected $p$-values for every contrast are in the released
    \texttt{ladder\_analysis.json}.}
  \label{tab:ladder}
  \setlength{\tabcolsep}{5pt}
  \footnotesize
  \begin{tabular}{lrrrr}
    \toprule
    & MNIST & Fashion & CIFAR-10 & CIFAR-100 \\
    & 2$\times$1000 & 2$\times$1000 & 3$\times$2000 & 3$\times$2000 \\
    \midrule
    \multicolumn{5}{l}{\emph{Test accuracy against \BP{}} (pp)} \\
    \quad \BP{} & $+0.00$ & $+0.00$ & $+0.00$ & $+0.00$ \\
    \quad \BPDS{} & $+0.07$ & $+0.04$ & $-1.65$ & $-1.06$ \\
    \quad \MFJ{} & $+0.14$ & $-0.09$ & $-0.19$ & $+1.86$ \\
    \quad \MF{} & $+0.19$ & $+0.01$ & $+3.90$ & $+5.04$ \\
    \addlinespace[2pt]
    \multicolumn{5}{l}{\emph{GPU energy} (\% of \BP{})} \\
    \quad \BP{} & $100$ & $100$ & $100$ & $100$ \\
    \quad \BPDS{} & $94$ & $107$ & $83$ & $79$ \\
    \quad \MFJ{} & $109$ & $131$ & $92$ & $118$ \\
    \quad \MF{} & $364$ & $403$ & $320$ & $377$ \\
    \bottomrule
  \end{tabular}
\end{table}

\subsection{Own Convergence: Accuracy BP Does Not Reach}
\label{sec:convergence}

The matched budget of Section~\ref{sec:matched} stops \MF{} early, and lifting it shows
what the method is actually capable of. Allowed instead to train until it stops
improving, \MF{} \emph{surpasses} \BP{} by \num{3.90}~pp
$[\num{+3.68}, \num{+4.11}]$ on CIFAR-10 and \num{5.04}~pp
$[\num{+4.81}, \num{+5.24}]$ on CIFAR-100, both pre-registered \num{95}\%
bootstrap intervals over \num{20} paired seeds. On the two
\num{2}$\times$\num{1000} tasks it matches \BP{} on Fashion-MNIST and leads it
on MNIST, where the pre-registered contrast is DIFFERENT at $+\num{0.25}$~pp
$[\num{+0.13}, \num{+0.38}]$ (Table~\ref{tab:conv}). Under the matched
data-pass protocol \MF{}'s energy is \SIlist{98;96;80;88}{\percent} of \BP{}'s
(Table~\ref{tab:matched}); at its own convergence,
\SIrange{320}{403}{\percent} (Table~\ref{tab:conv}). Even at its own convergence \MF{} draws \SI{19.1}{\percent} and
\SI{14.3}{\percent} less board power than \BP{} on CIFAR-10 and CIFAR-100; the
extra cost is time, not power.

The two protocols do not contradict each other. Per data pass \MF{} is the cheaper algorithm, by the margins
of Table~\ref{tab:matched}. It then takes \SIrange{3.5}{4.5}{\times} as many
passes to reach its own stopping point, from the realised Protocol~A epoch
medians (\numrange{90}{207} \MF{} epochs against \BP{}'s
\numrange{26}{52}), and the product is
\SIrange{3.2}{4.0}{\times} \BP{}'s energy end to end. Wall-clock time grows by a similar factor,
\SIrange{3.6}{4.4}{\times}, consistent with the near-equal time per pass
measured under Protocol B. The per-pass advantage
remains; the additional passes outweigh it, and Section~\ref{sec:limits}
treats the multiplier as the open problem.

\subsection{Quantifying Two Objections}
\label{sec:objections}

\paragraph{Is the recomputation simply an unoptimised implementation?}
\MF{} recomputes the frozen prefix at every stage, an $O(L^2)$ cost that
caching should remove, so we implemented the cache. It is bit-exact on the two \num{2}$\times$\num{1000} MLPs: over \num{1200} and \num{3600} optimiser steps the loss
and weights match the recomputing path exactly. It saved no time: on-device caching moved wall-clock by
$+\SI{3.8}{\percent}$ and $+\SI{2.7}{\percent}$ and cost \SI{374}{\percent} and
\SI{404}{\percent} more per-process memory; host caching was worse on every
axis. The cache does not help, because at
these widths the device is launch-bound rather than arithmetic-bound. Only the
local method can cache at all: in \BP{}, \BPDS{} and \MFJ{} a joint gradient
makes a cached activation stale within one optimiser step.

\begin{table}[tb]
  \centering
  \caption{Own convergence (Protocol A). Ratios are \MF{} over \BP{}. $\Delta$pp is a difference of arm means; on
    MNIST, where $n$ differs between arms, the paired contrast is
    $+\num{0.25}$~pp. Seeds: $n=\num{20}$ per arm on the CIFAR
    configurations, $n=\num{60}/\num{20}$ and $n=\num{122}/\num{48}$ on MNIST
    and Fashion-MNIST.}
  \label{tab:conv}
  \setlength{\tabcolsep}{8pt}
  \footnotesize
  \begin{tabular}{lrrrrr}
    \toprule
    Configuration & \BP{} acc. & \MF{} acc. & $\Delta$ & Time & Energy \\
                  & (\%)       & (\%)       & (pp)     & $\times$ & $\times$ \\
    \midrule
    MNIST 2$\times$1000 & 98.20 & 98.40 & +0.19 & 3.64 & 3.64 \\
    Fashion 2$\times$1000 & 89.41 & 89.42 & +0.01 & 4.21 & 4.03 \\
    CIFAR-10 3$\times$2000 & 62.47 & 66.37 & +3.90 & 3.96 & 3.20 \\
    CIFAR-100 3$\times$2000 & 32.90 & 37.94 & +5.04 & 4.39 & 3.77 \\
    \bottomrule
  \end{tabular}
\end{table}

\paragraph{Does \BP{} benefit from more mature libraries?}
If kernel maturity decided the comparison, \BP{} would run near saturation,
unlike the forward-only methods. Utilisation traces show otherwise: on the four ladder configurations no method exceeds
\SI{20.3}{\percent} mean utilisation, and across the entire study, the wider \FF{} MLPs included, no run exceeds
\SI{38.0}{\percent} mean utilisation, still far from saturation. Every arm is launch-bound, so \BP{} does not win by
saturating the device with more efficient arithmetic. This does not entirely
exclude an advantage from implementation maturity, but in these measurements
the end-to-end difference between \MF{} and \BP{} lies in the number of passes,
not in the cost of a pass.

\subsection{The Other Two Algorithms}
\label{sec:ffcafo}

\FF{} and \CaFo{} are the rest of the forward-only landscape, and neither
reproduces \MF{}'s combination of a per-pass cost advantage with competitive
accuracy. \FF{} trails \BP{} in accuracy on its native MLPs by
\num{0.01} to \num{1.10}~pp across its four arms, and requires
\SIrange{4.1}{7.4}{\times} \BP{}'s wall-clock time and
\SIrange{4.1}{6.3}{\times} its energy, with memory at \SIrange{0.91}{1.06}{\times} \BP{}'s
allocator peak. Removing the backward pass therefore does not by itself
produce the saving.
\CaFo{} costs \SIrange{3.0}{7.1}{\times} \BP{}'s energy. It lands within
\num{0.4}~pp of \BP{} on the two MNIST-scale convolutional neural networks (CNNs) but loses \num{18.8} and
\num{16.9}~pp on CIFAR-10 and CIFAR-100; its direct-feedback-alignment variant
narrows that to \num{7.1} and \num{8.5}~pp at a further
\SIrange{1.5}{2.6}{\times} the energy. Forward-only training is not
generically cheaper: the advantage measured in this paper belongs to
gradient locality as \MF{} implements it, not to the family.
Seeds are five per arm for the CNNs and seven per arm for the MLPs.

\section{Discussion and Limitations}
\label{sec:discussion}

\paragraph{Why gradient detachment might improve accuracy.}
The ladder shows a mechanism rather than just a performance gap: auxiliary
supervision alone makes \BP{} worse on the CIFAR tasks by
\num{1.06}--\num{1.65}~pp, precisely where \MF{} gains
\num{3.90}--\num{5.04}~pp. This means the improvement comes from detaching the
gradient, not from the local objective. One
interpretation is that greedy layer-wise optimisation constrains each layer's
effective capacity and so generalises better than joint
optimisation~\cite{ren2023scaling,ishikawa2025local}; four configurations and
one architecture family cannot establish this, so we state it as a hypothesis.

\paragraph{The open problem: convergence rate, not per-unit cost.}
\label{sec:limits}
\looseness=-1
Among the factors this study measures, one stands between \MF{} and an
end-to-end energy advantage. \MF{} needs \SIrange{3.5}{4.5}{\times} \BP{}'s data passes
and \SIrange{3.6}{4.4}{\times} its wall-clock time to reach its own best
accuracy, which converts a per-pass advantage of \SIrange{12.5}{19.9}{\percent}
into an end-to-end cost of \SIrange{3.2}{4.0}{\times}. The measurements
argue against the alternatives: the cost of a pass favours \MF{}; the cache
that removes the recomputation is bit-exact and saves no time; and every
ladder arm is launch-bound below \SI{21}{\percent} utilisation
(Section~\ref{sec:objections}), which makes faster arithmetic an unlikely
remedy, although it does not entirely exclude an implementation-maturity
effect. What remains is the optimisation schedule: the number of passes a
layer-sequential method needs before each stage converges, which is a
tractable target.

Three statements bound the result. \emph{Scope}: the
\SIrange{12.5}{19.9}{\percent} energy and \SIrange{24.6}{27.5}{\percent} memory figures
are measured on the \num{3}$\times$\num{2000} MLPs; at \num{1.8}~M weights the
memory advantage has not yet appeared, and we ran no convolutional \MF{}.
Flattening CIFAR for the MLP protocol caps accuracy far below the
\SI{85.6}{\percent} our own \num{3}-block CNN reaches under \BP{}, so these
comparisons are internally valid and say nothing about a competitive vision
pipeline.
\emph{Accuracy under the budget}: equivalence to \BP{} is established for CIFAR-100 under
the matched data-pass protocol, and the budget costs \MF{} \num{1.11}~pp on CIFAR-10,
so the matched data-pass comparison should be read as a cost result.
\emph{Hardware}: one accelerator, one driver. Board power at
this utilisation is dominated by the static and clock-domain components of a
large A100 die, so a smaller or better-saturated device would likely narrow the
power gap; the memory saving and the pass multiplier, which follow from
activation lifetime and from the optimisation, should carry across.

\paragraph{Implications.}
For a practitioner training \num{3}$\times$\num{2000} MLPs under a fixed
pass budget on this hardware, \MF{} is the cheaper choice: lower on
energy, power and memory, more so in the larger configurations, and on CIFAR-100 at no
accuracy cost. For anyone
training to convergence it reaches accuracy that \BP{} does not, at
\SIrange{3.2}{4.0}{\times} \BP{}'s energy. The methodological
point generalises: an efficiency claim about a training algorithm is
uninterpretable without its stopping rule, which is a term in it, not a
detail.

\section{Conclusion}
\label{sec:conclusion}

We measured the cost of forward-only training on hardware over almost
\num{1000} runs. Given the same number of data passes as \BP{}, Mono-Forward uses
\SIrange{12.5}{19.9}{\percent} less energy and \SIrange{24.6}{27.5}{\percent}
less per-process memory on the \num{3}$\times$\num{2000} MLPs, at wall-clock
time within \SI{3}{\percent}, with accuracy statistically equivalent to
\BP{}'s on CIFAR-100; trained to its own convergence it exceeds \BP{}'s test
accuracy by \num{3.90} and \num{5.04}~pp on the two CIFAR tasks. A
pre-registered ablation ladder identifies gradient detachment, not the
localised objective, as the dominant factor tested. The pass multiplier
separates per-pass from end-to-end advantage.

Future work should first reduce the number of passes \MF{} needs, for example
through per-stage learning-rate schedules and adaptive per-stage stopping criteria. Convolutional
and deeper architectures, training sets of varying size for a fixed task, and
other accelerators would then show how far the result extends. Importantly, the results matter most at scale:
activation memory and energy are major costs of training large models,
including large language models, and the saving we measure is larger in the
larger configurations tested. Whether it persists at that scale is untested
here, and establishing it is the natural next step.

\section*{Reproducibility}
\label{sec:repository}

Code, configurations and the tidy table of \num{19944} measurements behind
every number in this paper are available at
\url{https://github.com/Przemyslaw11/BeyondBackpropagation}; the per-run JSON
records and \num{1027} NVML power traces are archived and available from the
authors. The released analysis scripts regenerate every number, table body and
figure above.

\begin{credits}
\subsubsection{Author Contributions.}
The research was conducted as part of a master's thesis by
\mbox{Przemys{\l}aw} Spyra, under the supervision of Witold Dzwinel.
\mbox{Przemys{\l}aw} Spyra was responsible for the development and
implementation of the methodology, experiments, and analysis. Witold Dzwinel
provided conceptual guidance, critical feedback, and helped shape the direction
and presentation of the work.

\subsubsection{\ackname}
This work was supported by the National Science Centre (NCN), Poland, under
grant OPUS-29, DEC-2025/57/B/ST6/04377, and by the Ministry of Science and
Higher Education of Poland. The authors wish to thank the PLGrid Infrastructure
for its support and ACK Cyfronet AGH for providing the computational resources
on the Athena cluster under grant PLG/2025/018341, which were indispensable for
this work. We gratefully acknowledge AGH University of Krakow for the
institutional support and academic environment that made this research
possible. We also thank the reviewers, whose questions about the deep
supervision control and implementation maturity directed the two experiments
that most changed this paper.

\subsubsection{\discintname}
The authors have no competing interests to declare.
\end{credits}

\bibliographystyle{splncs04}
\bibliography{CP025}

\end{document}